\pdfoutput=1

\documentclass[11pt]{article}

\usepackage{emnlp2021}

\usepackage{times}
\usepackage{latexsym}

\usepackage[T1]{fontenc}

\usepackage[utf8]{inputenc}

\usepackage{microtype}

\usepackage{float}
 \usepackage{amsmath}
\usepackage{amsfonts}
\usepackage{verbatim}
\usepackage{multirow}
\usepackage{multicol}
\usepackage{graphicx}
\usepackage{tabularx}
\usepackage{relsize}
\usepackage{booktabs}
\usepackage{url}
\usepackage{xspace}
\usepackage{cleveref}
\usepackage{adjustbox}

\newcommand{\tabref}[1]{Table~\ref{#1}\xspace}
\newcommand{\figref}[1]{Figure~\ref{#1}\xspace}
\newcommand{\eqnref}[1]{Eqn~(\ref{#1})\xspace}
\newcommand{\secref}[1]{Section\xspace\ref{#1}\xspace}

\newcommand{\similarity}{\ensuremath{\operatorname{sim}}}
\newcommand{\z}{\phantom{0}}

\title{KFCNet: Knowledge Filtering and Contrastive Learning Network for Generative Commonsense Reasoning}

\author{Haonan Li$^1$\thanks{\ \ Work done during an internship at Microsoft Research Asia.} , 
        Yeyun Gong$^2$\thanks{\ \ Corresponding author.}, Jian Jiao$^3$, Ruofei Zhang$^3$, Timothy Baldwin$^1$, Nan Duan$^2$ \\
        $^1$School of Computing and Information Systems, The University of Melbourne \\
        $^2$Microsoft Research Asia $^3$Microsoft \\
        \url{haonanl5@student.unimelb.edu.au}, \url{tb@ldwin.net} \\
        \url{{yegong,Jian.Jiao,nanduan,bzhang}@microsoft.com}}

\begin{document}

\maketitle

\begin{abstract}
Pre-trained language models have led to substantial gains over a broad range of natural language processing (NLP) tasks, but have been shown to have limitations for natural language generation tasks with high-quality requirements on the output, such as commonsense generation and ad keyword generation. 
In this work, we present a novel \textbf{K}nowledge \textbf{F}iltering and \textbf{C}ontrastive learning \textbf{Net}work (KFCNet) which references external knowledge and achieves better generation performance. Specifically, we propose a BERT-based filter model to remove low-quality candidates, and apply contrastive learning separately to each of the encoder and decoder, within a general encoder--decoder architecture. 
The encoder contrastive module helps to capture global target semantics during encoding, and the decoder contrastive module enhances the utility of retrieved prototypes while learning general features.
Extensive experiments on the CommonGen benchmark show that our model outperforms the previous state of the art by a large margin: +6.6 points (42.5 vs.\ 35.9) for BLEU-4, +3.7 points (33.3 vs.\ 29.6) for SPICE, and +1.3 points (18.3 vs.\ 17.0) for CIDEr.
We further verify the effectiveness of the proposed contrastive module on ad keyword generation, and show that our model has potential commercial value.

\end{abstract}

\section{Introduction}

Pre-trained language models have achieved impressive results across a wide range of NLP tasks \citep{BERT, XLNet, ERNIE, RoBERTa, BART, ProphetNet, DeBERTa}. However, their ability to accurately reflect factual knowledge or perform logical inference is still limited.
To investigate the ability of systems to capture commonsense knowledge, datasets such as CommonsenseQA \citep{CommonsenseQA}, SWAG \citep{SWAG}, and WinoGrande \citep{WinoGrande} have been proposed.
Separate to these discriminative tasks that require models to choose the correct option from multiple candidates, CommonGen \citep{CommonGen} is framed as a generation task, and requires the system to generate a logical and coherent sentence describing an everyday scenario based on a concept set.
Experiments show that state-of-the-art generation models are not adequate or accurate enough to generate plausible sentences or reflect commonsense assumptions in this setting.

External knowledge provides not only information about the sorts of relationships that hold between concepts, to potentially guide generation models in capturing the implicit logic between concepts, but also interpretability.
Inspired by \citet{RAG} and \citet{EKI}, we adopt a retrieval-and-generation framework, and propose a BERT-filter and two contrastive learning modules for retrieval and generation, respectively.

For retrieval, previous research \citep{RAG} has shown that traditional sparse vector space models, such as TF-IDF and BM25, perform better than dense representation-based retrieval on heavily entity-centric tasks such as FEVER \citep{FEVER}.
However, while using sparse vector space retrieval models can retrieve relevant prototypes that contain a set of concepts, there can be significant domain mismatches between the retrieved results and target distribution, making it difficult for generation models to bridge between prototypes and targets.
We argue that a two-stage retrieval strategy alleviates this issue by combining sparse vector space search and dense representation filters.
First, a sparse vector retrieval model is used to find passage candidates with high coverage of concept words, and then a dense vector-based filter is applied to score the candidates, and filter out low-quality prototypes.

For generation, we apply contrastive learning to each of the encoder and decoder, in a general encoder--decoder architecture. 
The core idea of contrastive learning is to construct positive and negative samples from an anchor sample, and draw together the anchor and positive samples while pushing away the anchor from all negative samples in the embedding space during training. 
Given that high-quality prototypes can be used as clusters of positive samples, we propose a decoder contrastive module that minimizes the distance between decoded sentence representations with distinct prototypes retrieved from the same concept set. 
Common scenario information and abstract concept relationships can be learned based on the contrasts between different prototypes.
Moreover, we propose an encoder contrastive module to force the encoder to learn sentence representations, and save it into a global token which is visible to the decoder during decoding. In this way, global sentence-level semantics can be captured better.

The main contributions of this work are threefold:
(1) we demonstrate that adding a high-quality matching model to the word overlap-based retriever benefits entity-centric retrieval tasks;
(2) we propose two contrastive learning modules that can be applied to a general encoder--decoder generation model; and
(3) we conduct experiments on CommonGen and an ad keyword generation task, and show that our method achieves large-scale improvements on both tasks.


\section{Related Work}
\label{sec:related}

\subsection{Knowledge Enhanced Generation}

There is significant work on incorporating external knowledge from knowledge bases and incorporating retrieved information in language generation tasks \citep{DBLP:conf/emnlp/WestonDM18, DBLP:conf/acl/LiWLC18, DBLP:conf/aaai/GuanWH19, DBLP:conf/acl/HossainGZ20}.
\citet{RAG} explore a general-purpose fine-tuning recipe for retrieval-augmented generation that combines a dense passage retriever \citep{DPR} with a BART \citep{BART} generator. 
For commonsense generation, \citet{KG-BART} propose a knowledge graph-augmented language generation model that encompasses concepts from a knowledge graph, and produces more logical and natural sentences.
\citet{EKI} propose to retrieve prototypes based on sparse vector similarity, and introduce a scaling module and a prototype position indicator to explicitly deal with retrieval noise. 

This paper proposes a two-stage retrieval strategy and differs in applying contrastive learning to make better use of prototypes for generation.

\subsection{Contrastive Learning}

Recently, contrastive learning has achieved remarkable results in many self-supervised and supervised learning tasks, primarily for computer vision.
The two key elements of contrastive learning are: (1) the construction of positive and negative samples; and (2) the learning framework.

\subsubsection{Sample Construction}
Usually in contrastive learning, positive samples are augmented forms of anchor data points, and negative samples are augmented forms of other data points.
In NLP, \citet{Coco-LM} create positive samples by masking and cropping tokens from sentences;
\citet{DBLP:journals/corr/abs-2011-01403} and \citet{CERT} use back-translation to create positive augmentations of original sentences;
\citet{InfoXLM} and \citet{HICTL} regard parallel sentences distributed in one or multiple languages as different views of the same semantics to learn cross-lingual representations;
and \citet{SimCSE} demonstrate that constructing positive pairs with only standard dropout as minimal
data augmentation works surprisingly well on the NLI task.
Distinct from these methods, we propose to create positive sample pairs from retrieved results.

\subsubsection{Learning Framework}
Previous contrastive learning methods have required either specialized architectures \citep{DBLP:conf/nips/BachmanHB19, CPC} or a memory bank to store large volumes of negative samples \citep{DBLP:conf/cvpr/WuXYL18, DBLP:conf/eccv/TianKI20}. 
\citet{SimCLR} present a simple framework consisting of a feature extraction module, and a non-linear transformation module, which outperforms previous work on ImageNet \citep{ImageNet} without using a specialized architecture or a memory bank.
However, it requires a large batch size to yield high performance, which is computationally prohibitive.
Moco \citep{Moco} addresses this issue by maintaining a queue of data samples as the memory bank, and enqueuing encoded representations of the current mini-batch and dequeuing the oldest representations on each iteration. 
They further propose a momentum encoder to maintain the consistency of representations in the queue.
In this work, we use the Moco framework to train our contrastive learning modules.

\section{Method}
\label{sec:method}
In this section, we detail our method: \textbf{K}nowledge \textbf{F}iltering and \textbf{C}ontrastive learning \textbf{Net}work (KFCNet).
First, we introduce our prototype retrieval strategy together with the knowledge filter model. 
Then we present our generative model, based on an encoder--decoder architecture with two contrastive learning modules.
Finally, we show how we adapt the Moco contrastive learning framework, and deal with multiple positive samples.

\subsection{Task Formulation}

We use $S$ to denote a set of concepts, where $S=\{c_1, c_2,..., c_m\}$,  $c_i \in \mathcal{C}$ and $\mathcal{C}$ is the concept vocabulary, and use $\mathcal{X}$ to denote all possible concept sets. 
The commonsense generation task is to learn a function $f:\mathcal{X}\rightarrow\mathcal{Y}$ that maps the concept set $S$ to a sentence $T$, where $T=\{t_1, t_2,..., t_n\} \in\mathcal{Y}$, and $\mathcal{Y}$ is the target sentence space. The generated sentence must be a plausible sentence that describes a common scenario in our daily life based on the contents of $S$.

\subsection{Prototype Retrieval and Filtering}

In order to retrieve prototypes that contain the concepts in a given concept set while keeping the retrieval results and target sentences semantically as similar as possible, we use a two-stage retrieval strategy combining sparse vector space search and dense representation matching. 
In Stage 1, a sparse vector retrieval model is used to retrieve $N$ candidate prototypes from the corpus $\mathcal{D}$, where $N \ll \left | \mathcal{D} \right |$.  
Then in Stage 2, a dense representation-based scorer is used to score the candidates, and the top-$k$ scored candidates are chosen as the final prototypes.

\subsubsection{Stage 1} 
Given a concept set $S=\{c_1, c_2, ..., c_m\}$, we first split corpus $\mathcal{D}$ into $m+1$ parts $\{d_0,d_1,..., d_{m}\}$ according to the number of concepts the sentence contains, where sentences in $d_i$ contain $i$ concepts in $x$. 
Given that most concepts in $\mathcal{C}$ are verb and noun lemmas, we pre-process based on lemmatization and stemming. 
Then we choose $N$ sentences as candidates from the parts, prioritized such that $d_m > d_{m-1} ... > d_1$.

\subsubsection{Stage 2} 
After retrieving $N$ candidate prototypes, a scorer is used to rank the candidates and filter out candidates that are far from the targets in embedding space. 
In this work, we use a BERT-based model as an encoder, and use the hidden state of the \texttt{[CLS]} token as the sample representation. 
The representation is then feed into a multi-layer perceptron with a scalar output as follows: 
\begin{align}
    r_{cls} &= \operatorname{BERT}(S) \\
    score &= \operatorname{MLP}(r_{cls})
\end{align}
where $S = \texttt{[CLS]} + concept\ set + \texttt{[SEP]} + sentence + \texttt{[SEP]}$ is the training sample created from the concept set and candidates/targets, $r_{cls} \in \mathbb{R}^{d}$ is the sample representation, and $score \in [0,1]$ is the final score of the sample.
Theoretically, the label of the training set can be any real number in the range $[0,1]$, but we find that it is sufficient to train the scorer as a simple binary classifier.
We create the positive samples by combining a concept set with each corresponding target sentence, and create the negative samples by combining the concept set with a different candidate prototype or random sentence from $\mathcal{D}$.
During inference, we score all candidates and choose the candidate with the highest score as the prototype.

For all experiments in this paper, we set $k=2$ and $N=8$.
$k=2$ means that for each input, we construct one positive sample which is widely used in contrastive leaning work.
$N=8$ is because experience shows that at least 2 high-quality prototypes can be retrieved with 8 candidates.

\begin{figure*}[t]
	\centering
	\includegraphics[width=0.85\textwidth]{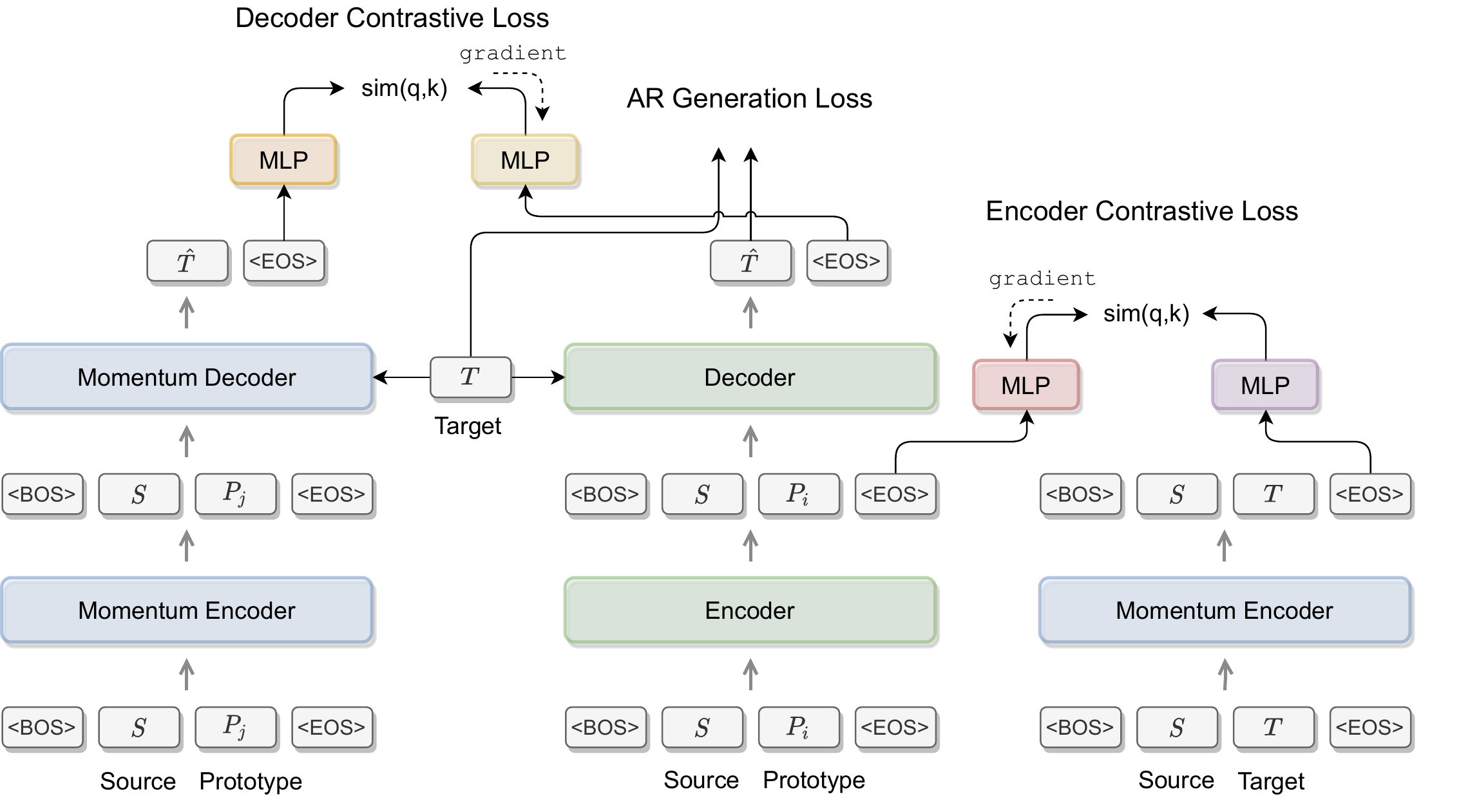}
	\caption{Generation model structure with contrastive learning.}
	\label{fig:model}
\end{figure*}

\subsection{Contrastive Learning for Generation}

\subsubsection{Encoder--decoder Architecture}
The encoder--decoder architecture is widely used in generation tasks. Compared to single decoder generation models such as GPT-2 \citep{GPT} where words are conditioned only on left context, models using an encoder--decoder framework such as BART \citep{BART} and T5 \citep{T5} enable bidirectional interactions with an encoder, and auto-regressive generation with a decoder.
In this work, we use BART with an auto-regressive objective for generation as shown in the middle of Figure \ref{fig:model}, and propose separate contrastive modules for the encoder and decoder, corresponding to the right and left sub-structures in the figure.

Typical generation inputs only contain the source sentence, which is the concept set $S$ in CommonGen. The difference here is that we append one of the retrieved prototypes $p_{i} \in P$ to $S$ to enhance the input.

\subsubsection{Encoder Contrastive Loss}
Although BART learns bidirectional interactions using a transformer-based encoder and implements cross-attention at each layer of the decoder, the global target information is not explicitly learned during encoding, meaning that the decoder needs to find important local information for the current step during each timestep of decoding, without having access to the global goal of generation.
Here, we propose to force the model to learn global target information during encoding and save it to a special token, using source--target contrastive learning. 
The special token is visible to the decoder via cross-attention during each timestep of decoding.
Specifically, given a concept set $S=\{c_1, c_2,..., c_m\}$ and a target sentence $T=\{t_1, t_2,..., t_n\}$, where $c_i \in \mathcal{C}$, we denote the retrieved prototypes as $\{p_1, p_2,..., p_k\}$, where each $p_i$ is a complete prototype.
We construct the input for encoder contrastive learning by concatenating $S$ with $T$ and $S$ with $p_i$, respectively.
As illustrated in \figref{fig:model}, the concatenation of $S$ and $p_i$ will be used as the input to the main encoder, which is followed by a decoder with gradient and auto-regressive generation loss, and the concatenation of $S$ and $p_i$ will be used as the input to another encoder without a decoder or gradient. 

\subsubsection{Decoder Contrastive Loss}
At the retrieval step, multiple high-quality retrieval results are collected as prototypes to augment generation. Although these retrieved results substantially boost external information, they inevitably introduce noise. In order to learn general information associated with the concept set and eliminate noise in the prototypes, we propose a decoder contrastive learning module, which we apply to the sentence representation at the decoder output.
Formally, we concatenate $S$ with $p_j$ ($j\neq i$), which is a different prototype for $S$ from the one used in the main-branch BART model. Note also that different from the main-branch model, here the gradient is not back-propagated.

\subsection{Momentum Contrast with Memory Bank}

Most existing training methods greatly limit the number of in-batch negative samples, limiting the potential of contrastive learning.
To enable large-scale interactions between negative samples, we follow Moco \citep{Moco} in maintaining a dictionary as a queue of encoded/decoded data samples.
The keys of the dictionary are samples from data after encoding/decoding and the queries are samples in current mini-batches after encoding/decoding during training.
Learning is formulated as minimizing the contrastive loss, which makes the query similar with its matching key and dissimilar to other keys.

\subsubsection{Memory Bank as Queues} 
We use two dictionaries to store the representations of the encoder and decoder output, respectively. 
In each training iteration, the newest encoded representations are enqueued and the oldest are dequeued, to maintain a fixed queue size. 
For each sample, the number of contrast pairs is the size of the queue, where usually only the matching key in the same mini-batch is positive, and all others are negative.

\subsubsection{Dealing with Multiple Positive Samples}
In the CommonGen task, the mapping between source sequences and gold targets can be many-to-many.
$N$ independent sample pairs are created for 1-to-$N$ and $N$-to-1 source--target pairs, which can be distributed across mini-batches. 
However, these $N$ samples should be all regarded as mutually positive. 
To enable interactions between positive samples intra- and inter-mini-batch, we assign each source--target pair an identity, which indicates pairs that share the same source or target. 
These identities are saved in another queue that is synchronously updated with an encoder and decoder memory bank.

\subsubsection{Momentum Updated Parameters}
To keep the consistency of representations in the memory banks, we update the parameters of the key encoder and decoder with momentum.
Formally, denote the parameters of the query encoder and decoder as ${\theta}_{q}^{e}$ and ${\theta}_{q}^{d}$, and the parameters of the key encoder and decoder as ${\theta}_{k}^{e}$ and ${\theta}_{k}^{d}$. 
The parameters ${\theta}_{q}^{e}$ and ${\theta}_{q}^{d}$ are updated by back-propagation, and the parameters  ${\theta}_{k}^{e}$ and ${\theta}_{k}^{d}$ are updated by:
\begin{align}\label{eqn:momentum1}
    {\theta}_{k}^{e} \leftarrow  m{\theta}_{k}^{e} + (1-m){\theta}_{q}^{e} \\
    {\theta}_{k}^{d} \leftarrow  m{\theta}_{k}^{d} + (1-m){\theta}_{q}^{d}
\end{align}
Here, $m\in [0,1)$ is a momentum coefficient which is set to be close to 1. In this way, the parameters of the key encoder and decoder evolve more smoothly than those of the query, which maintains the consistency of key representations in the memory bank.

\subsection{Training Objective}
\label{sec:objective}

Consider a batch of query-key pairs $\{(q_1, k_1), (q_2, k_2), ...,  (q_n, k_n)\}$, where there is only one positive key $k_i$ for a given query $q_i$. 
After encoding, we fetch the representation of the last \texttt{<EOS>} tokens and apply a projection to it as:
\begin{align}
    a_i^{eos} &= \operatorname{Encoder}(q_i) \\
    z_i &= \operatorname{Proj}(a_i^{eos})
\end{align}
The encoder contrastive loss function, called InfoNCE, is as follows:
\begin{equation}
    \mathcal{L}_{E_i}  = - \log \frac{\exp(\similarity({z_{q_i}, z_{k_i}}) /\tau )}{\sum_{j \in M} \exp(\similarity({z_{q_i}, z_{k_{j}}) /\tau })} \label{eq:infonce}
\end{equation}
where $\similarity(,)$ denotes the similarity function, $\tau$ is a temperature hyper-parameter, and $M$ denotes all indices in the memory bank.
The denominator has $|M|$ total terms, including one positive and $|M|-1$ negative samples.
Intuitively, the loss function is the log loss of an $|M|$-way softmax classifier that tries to classify $q_i$ according to the positive $k_i$.
\eqnref{eq:infonce} is only able to deal with the case of a single positive key existing for each query. 
To generalize it to an arbitrary number of positives, inspired by SupCon \citep{SupCon}, we consider the following loss functions:
\begin{align}
    \mathcal{L}_{E_i}^{out} &= - \sum_{p \in P(i)} \frac{1}{|P(i)|} \log \mathcal{L}_{E_{i,p}}^{single} \label{eq:out} \\
    \mathcal{L}_{E_i}^{in}  &= - \log \left \{\frac{1}{|P(i)|} \sum_{p \in P(i)} \mathcal{L}_{E_{i,p}}^{single} \right \} \label{eq:in} 
\end{align}
\begin{align}
    \mathcal{L}_{E_{i,p}}^{single} &= \frac{\exp(\similarity({z_{q_i}, z_{k_p}}) /\tau )}{\sum_{j \in M} \exp(\similarity({z_{q_i}, z_{k_{j}}) /\tau })}
\end{align}
Here, $P(i)$ denotes all positive indices of the sample $i$, \eqnref{eq:out} summarizes the positive samples outside of the log function, and \eqnref{eq:in} summarizes those inside it.

The decoder contrastive loss $\mathcal{L}_{D}$ can be obtained in the same way, except that the sentence representation is fetched from the \texttt{<EOS>} token after decoding. During training, we try to minimize the sum of the encoder contrastive loss, decoder contrastive loss, and the decoder auto-regressive generation loss:
\begin{equation}
    \mathcal{L} = \mathcal{L}_{CE} + \lambda_1 \mathcal{L}_E  +\lambda_2 \mathcal{L}_D
\end{equation}
Here, $\mathcal{L}_{CE}$ denotes the cross-entropy generation loss, and $\lambda_1$ and $\lambda_2$ are tunable scalars.

During inference, we discard the momentum encoder and decoder, together with the projection layers.

\section{Experiments}
\label{sec:experiment}

\begin{table}
	\centering
          \begin{adjustbox}{max width=\columnwidth}
	\begin{tabular}{lcc}
		\toprule
		& \textbf{$\mathcal{D}_{in}$} & \textbf{$\mathcal{D}_{out}$} \\
		\midrule
		\# Sentences & 4,118,993 & 70,245,048 \\
		\# Length (avg) & 11.10 & 18.76 \\
		\# Missing concept (avg) & & \\
		\qquad size=3 & 0.40  & 0.48 \\
		\qquad size=4 & 0.56  & 1.00 \\
		\qquad size=5 & 0.80  & 1.98 \\
		\bottomrule
	\end{tabular}
        \end{adjustbox}
          \caption{Statistics of the two corpora. ``Missing concept''
          indicates the number of missing concepts in the top-2 retrieved sentences, broken down by concept-set size.}
	\label{tab:corpus}
\end{table}

\subsection{Datasets}
CommomGen \citep{CommonGen} contains 32,651/993/1,497  unique training/development/test concept sets, corresponding to 67,389 and 4,018 English target sentences in the training and development sets, meaning that one concept-set can map to multiple target sentences. The percentage of concept-sets in the development and test sets that are unseen in the training set are 99.60\%, and 100.00\% respectively, making the dataset challenging for compositional generalization.

\subsection{Prototype Collection}

\subsubsection{In-domain Corpus}

As CommonGen was created from visually-grounded caption datasets that describe everyday scenarios, we build an in-domain corpus from datasets of image captions, video captions, and natural language inference.
In detail, we extracted sentences from ActivityNet \citep{ActivityNet}, VaTeX \citep{VaTeX}, Conceptual Captions \citep{Conceptual}, SNLI \citep{SNLI}, and MNLI \citep{MNLI} as the in-domain corpus ($\mathcal{D}_{in}$).

\subsubsection{Out-of-domain Corpus} In addition to in-domain experiments, we create an out-of-domain corpus ($\mathcal{D}_{out}$) from Wikipedia,\footnote{English Wikipedia dump from May 01, 2020.}
using SpaCy\footnote{\url{https://spacy.io/}} as our sentence tokenizer.

For both corpora, sentences with fewer than 5 tokens or more than 100 tokens were removed. \tabref{tab:corpus} shows the basic statistics of the two corpora. 
Although $\mathcal{D}_{out}$ is much larger than $\mathcal{D}_{in}$, sentences retrieved from $\mathcal{D}_{in}$ contain more required concepts than those from $\mathcal{D}_{out}$ on average. 
Specifically, for concept-sets of size 4 and 5, the retrieved sentences from $\mathcal{D}_{out}$ have 0.44 and 1.18, respectively, more relevant concepts than $\mathcal{D}_{in}$.

\subsection{Experimental Setup}
\subsubsection{Implementation Details}
We employ the pre-trained BART-large model as the base generation model, and initialize the momentum encoder and decoder by copying parameters from the base model.
We use the Adam optimizer with ${\beta}_{1,2}=(0.9, 0.999)$, $\epsilon=1e-6$, and 0.1 weight decay, with the initial learning rate setting selected from $\{8e-6, 1e-5, 3e-5, 5e-5\}$. 
We use the polynomial decay learning rate scheduler with 500 warmup steps, and set dropout to 0.1.
We set the max tokens per batch to 3000 and max batch-size to 48, with 15k total updates.
For the auto-regressive generation loss, we use cross-entropy loss with 0.1 label-smoothing penalty.
During decoding, we use beam search with size 5, and 1.0 length penalty. 

For contrastive learning, we use an MLP as the projection network, with a single hidden layer of 1024d and the output size of 128d. 
We use \eqnref{eq:out} as the loss function, with similarity measured by dot-product, and set the temperature to 0.1.
The queue size of the memory bank is set to 4096, and the momentum coefficient is set to 0.999.

\subsubsection{Baselines}
We use several state-of-the-art pre-trained language generation models as baselines: GPT-2 \citep{GPT}, BERT-Gen \citep{UniLM-v2}, UniLM \citep{UniLM}, UniLM-v2 \citep{UniLM-v2}, T5 \citep{T5}, and BART \citep{BART}. 
All models are fine-tuned in seq2seq mode. 
We also compare our model with two strong baselines that use external knowledge: EKI \citep{EKI} and KG-BART \citep{KG-BART}.

\subsubsection{Evaluation Metrics} 

To evaluate generation performance, we use BLEU \citep{BLEU}, ROUGE \citep{ROUGE}, and METEOR \citep{METEOR}, in addition to evaluation metrics for captioning tasks, namely CIDEr \citep{CIDEr} and SPICE \citep{SPICE}. 
As all metrics score the output in the range $[0,100]$, we also present the average score across all metrics.

\section{Results}
\label{sec:results}
\begin{table*}[t!]
	\begin{center}
          \begin{adjustbox}{max width=\textwidth}
		\begin{tabular}{lcccccccc}
			\toprule
			\textbf{Model} & \multicolumn{2}{c}{\textbf{ROUGE-2/L}} & \multicolumn{2}{c}{\textbf{BLEU-3/4}} &\textbf{METEOR} & \textbf{CIDEr} & \textbf{SPICE} & \textbf{Overall}\\ 
			\midrule
            GPT-2 \citep{GPT}       & 17.18 & 39.28 & 30.70 & 21.10 & 26.20 & 12.15 & 25.90  & 24.64 \\
			BERT-Gen \citep{UniLM-v2}    & 18.05 & 40.49 & 30.40 & 21.10 & 27.30 & 12.49 & 27.30  & 25.30 \\
			UniLM \citep{UniLM}       & 21.48 & 43.87 & 38.30 & 27.70 & 29.70 & 14.85 & 30.20  & 29.44 \\
			UniLM-v2 \citep{UniLM-v2}    & 18.24 & 40.62 & 31.30 & 22.10 & 28.10 & 13.10 & 28.10  & 25.93 \\
			T5 \citep{T5}          & 22.01 & 42.97 & 39.00 & 28.60 & 30.10 & 14.96 & 31.60  & 29.89 \\
			BART \citep{BART}        & 22.23 & 41.98 & 36.30 & 26.30 & 30.90 & 13.92 & 30.60  & 28.89 \\
			\midrule
            EKI-out \citep{EKI}     & 24.36 & 45.42 & 42.90 & 32.10 & 32.00 & 16.80 & 32.50 & 32.29 \\
			KFCNet-out     & 24.10 & 45.59 & 44.09 & 34.20 & 32.83 & 17.39 & 33.11 & 33.04 \\
			\midrule
			KG-BART \citep{KG-BART}     & 23.38 & 44.54 & 42.10 & 30.90 & 32.40 & 16.83 & 32.70 & 31.83 \\
			EKI \citep{EKI}         & 25.43 & 46.53 & 46.00 & 36.10 & 33.80 & 17.80 & 33.40 & 34.15 \\
			\midrule
            KFCNet w/o FC  & 25.16 & 46.13 & 50.22 & 41.97 & 36.22 & 18.85 & 35.90 & 36.35 \\
            KFCNet w/o C   & 25.91 & 46.81 & 54.75 & 47.33 & 38.19 & 20.21 & 38.20 & 38.77 \\
            KFCNet         & \textbf{26.81} & \textbf{47.52} & \textbf{57.33} & \textbf{51.46} & \textbf{38.92}  & \textbf{20.98} & \textbf{39.15} & \textbf{40.31}\\
			\bottomrule
		\end{tabular}
                \end{adjustbox}
	\end{center}
	\caption{Overall performance of the different models on CommonGen (v1.0). 
    Models in the first block are fine-tuned pre-trained language models without external knowledge; models in the second block use out-of-domain knowledge; 
    models in the last two blocks use in-domain knowledge, where the KG-BART uses ConceptNet, and both EKI and KFCNet (our model) use the in-domain prototype corpus as a knowledge base.}
	\label{tab:main}
\end{table*}

\tabref{tab:main} presents the experimental results across all the metrics.\footnote{Note that the latest test set (v1.1) adds one more human reference to each example in the test set (v1.0), but is not publicly available. Additionally, EKI and KG-BART were evaluated on v1.0, so this is what we use for our experiments. }
We observe the following:
(1) Methods in the 2nd, 3rd, and 4th blocks of \tabref{tab:main} that use external knowledge outperform the fine-tuned pre-trained language models in the first block.
This demonstrates that external knowledge benefits commonsense reasoning and generation.
(2) The overall performance of EKI and our method (KFCNet) that both use natural sentences as prototypes is better than KG-BART, which incorporates structured knowledge from knowledge bases.
We hypothesize that this is because pre-trained language models like BART can more easily exploit natural language samples than structured information, even with elaborate modules for information fusion.
(3) Prototypes retrieved from the in-domain corpus result in better performance than those from the out-of-domain corpus.
(4) Simply fine-tuning BART on our retrieved prototypes beats previous published SOTA on several metrics, and using filtered prototypes boosts the performance again. 
This on the one hand shows that the quality of prototypes has a large impact on generation, and
on the other hand, indicates our retrieval method is better than that of EKI, and our filter helps in selecting good prototypes.
(5) Our KFCNet achieves new state-of-the-art results which surpass all other methods by a large margin.

\begin{table}
\centering
\begin{tabular}{lc@{\,\,}c@{\,\,}c}
\toprule
\textbf{Model} & \textbf{BLEU-4} & \textbf{CIDEr} & \textbf{SPICE} \\
\midrule
w/o Retrieval       & 26.30 & 13.92 & 30.60 \\
BM25                & 36.84 & 17.33 & 32.96 \\
\z$+$Lemma \& Stem      & 41.97 & 18.85 & 35.90 \\
\z$+$BERT Filter        & \textbf{47.33} & \textbf{20.21} & \textbf{38.20} \\
\bottomrule
\end{tabular}
\caption{Results for fine-tuning BART based on different retrieval strategies over the test set.}
\label{tab:retrieval}
\end{table}

\subsection{Ablation study}

To better understand the impact of the different modules in KFCNet, we perform a number of ablation experiments.

\subsubsection{Retrieval and Filter}
Prototype retrieval is a key part of any retrieval-based generation model. 
To assess the effectiveness of the retrieval-and-filter mechanism, we retrieve prototypes from the in-domain corpus and run ablations on a single BART model.
\tabref{tab:retrieval} shows the results.
Compared to models without retrieval, using prototypes retrieved by a simple BM25 model improves generation performance, which we suggest is due to the retrieved prototypes helping the model to better capture relationships between concepts, and construct a coherent scenario.
With word lemmatization and stemming, the variety of the retrieval results increases, resulting in better prototypes.
Adding a BERT filter boosts the performance again, achieving $+$5.38, $+$1.36, and $+$2.30 absolute improvements for BLEU-4, CIDEr, and SPICE. 
This verifies the effectiveness of using a high-quality matching model as an auxiliary module for a word overlap-based retriever.

\begin{table}
\centering
\begin{tabular}{lccc}
\toprule
\textbf{Model} & \textbf{BLEU-4} & \textbf{CIDEr} & \textbf{SPICE} \\
\midrule
KFCNet          & \textbf{36.10} & \textbf{17.96} & 33.89 \\
$-C_{D}$             & 34.09 & 17.24 & \textbf{33.97} \\
$-C_{E} -C_{D}$            & 30.82 & 16.20 & 33.16 \\
\bottomrule
\end{tabular}
\caption{Contrastive ablation study on CommonGen development set. $C_E$ and $C_D$ denote the encoder and decoder contrastive modules, respectively.}
\label{tab:contrastive}
\end{table}

\subsubsection{Contrastive Learning}
The contrastive loss plays an important role in our model. 
We perform an ablation study on the development set of CommonGen, by comparing the model without the contrastive module, using only encoder contrastive learning, and using both encoder and decoder contrastive learning.
As shown in \tabref{tab:contrastive}, using  only encoder contrastive learning leads to improvements over the baseline BART model, and adding decoder contrastive learning further improves results based on BLEU-4 and CIDEr.

\subsection{Similarity Function and Summation Location}

We further compare the performance of different similarity functions and positive summation locations, as mentioned in \secref{sec:objective}.
The results in \tabref{tab:sim_sum} demonstrate that the combination of dot-product similarity with summation outside of the log function performs best, consistent with the findings of \citet{SupCon}.

\begin{table}
\centering
\begin{tabular}{llccc}
\toprule
\textbf{Sim} & \textbf{Sum} & \textbf{BLEU-4} & \textbf{CIDEr} & \textbf{SPICE} \\
\midrule
    $(\cdot)$ & Out & \textbf{34.11} & \textbf{16.77} & 33.59 \\
    $(\cdot)$ & In & 32.45 & 16.54 & \textbf{33.79} \\
    $\cos(,)$ & Out & 32.49 & 16.62 & 33.73 \\
    $\cos(,)$ & In & 33.52 & 16.64 & 33.39 \\
\bottomrule
\end{tabular}
    \caption{Comparison of different similarity functions and positive sample summation locations. $(\cdot)$ denotes dot-product similarity, and $\cos(,)$ denotes cosine similarity.}
\label{tab:sim_sum}
\end{table}

\subsection{Model Efficiency}

\subsubsection{Retrieval}
The prototype retrieval is done separately from the generation model, and the retrieval time consists of 2 parts: 
(1) sparse vector matching time, in the form of BM25 search; and 
(2) BERT filter inference, for fine-grained selection, noting that only a few candidates (8 in our experiments) are left after stage 1, which can be processed in a single mini-batch.

\subsubsection{Contrastive Module} 
During training, the momentum encoder and decoder parameters are updated by \eqnref{eqn:momentum1} and there are no gradients or back-propagation in these modules. 
Therefore it takes no more than double the training time without contrastive modules.
During inference, the contrastive modules are disabled, and hence the efficiency does not decrease.

\subsection{Final Leaderboard Results }

\begin{table}
\centering
\begin{tabular}{lccc}
\toprule
\textbf{Model} & \textbf{BLEU-4} & \textbf{CIDEr} & \textbf{SPICE} \\
\midrule
Human           & 46.49	& 37.64 & 52.43 \\
\midrule
KFCNet 		    & \textbf{42.45}	& \textbf{18.37}	& \textbf{33.27} \\
RE-T5           & 40.86 & 17.66 & 31.07 \\
KG-BART 		& 33.86	& 16.92	& 29.63 \\
EKI-BART 		& 35.94	& 16.99	& 29.58 \\
T5-Large 		& 31.96	& 15.12	& 28.85 \\
BART 			& 31.82	& 13.97	& 27.99 \\
UniLM           & 30.61	& 14.88	& 27.42 \\
BERT-Gen 		& 23.46	& 12.60	& 24.82 \\
GPT-2           & 26.83	& 12.18	& 23.56 \\
\bottomrule
\end{tabular}
    \caption{Final CommonGen leaderboard results, using SPICE to rank the methods.}
\label{tab:leaderboard}
\end{table}

\tabref{tab:leaderboard} shows the final evaluation results on the latest test set with additional human references (v1.1).\footnote{\url{https://inklab.usc.edu/CommonGen/leaderboard.html}}
Note that the model in second place (RE-T5) expands the original training data and does continuous pretraining before fine-tuning on CommonGen. Our method, KFCNet, performs best on all metrics.
Among all fine-tuned methods, KFCNet beats the previous state-of-the-art by a large margin: $+$6.51 (18.11\%) for BLEU-4, $+$1.38 (8.12\%) for CIDEr, and $+$3.64 (11.95\%) for SPICE.

\subsection{Experiment on Keyword Generation}
To test the effectiveness of the proposed contrastive learning modules, we constructed a real-world adword dataset, based on an advertising platform \citep{edelman2007internet}.
The goal is to display a list of ads that matches the user intent, for which the first step is to retrieve relevant keywords provided by advertisers given a user query. 
The dataset contains 72,876 training samples, 10,000 dev samples, and 10,000 test samples from a major search engine, with each sample corresponding to a query--keyword pair.
Titles of the top-two web search results of the query from the search engine are kept as prototypes. 

We fine-tune BART models following the same sequence generation experimental design. The results are shown in \tabref{tab:commercial}.

From the first two lines, we see that directly appending the retrieved information to the source  does not lead to noticeable improvements, almost certainly because of noise in the retrieved results. However, our contrastive modules alleviate the effects of noise, and beat BART on all metrics.

\begin{table}
\centering
\begin{tabular}{lccc}
\toprule
\textbf{Model} & \textbf{ROUGE-2/L} & \textbf{BLEU-3/4} &\textbf{AVG} \\
\midrule
BART            & 33.03/60.17 & 31.61/25.03 & 38.96 \\
$+R$              & 33.68/60.31 & 31.69/24.75 & 39.70 \\
$+C_{E,D}$             & \textbf{35.18/61.24} & \textbf{33.60/26.78} & \textbf{41.44} \\
\bottomrule
\end{tabular}
\caption{Experimental results on ad keyword generation.}
\label{tab:commercial}
\end{table}

\section{Conclusion}
\label{sec:conclusion}

In this paper, we present KFCNet: a novel knowledge filtering and contrastive learning model for retrieval-augmented sequence generation, which achieves state-of-the-art results on the CommonGen benchmark.
Two contrastive learning modules are proposed to capture global target semantics and learn general features from multiple retrieved prototypes.
A prototype retrieval ablation study showed the effectiveness of the proposed filter for filtering low-quality candidates, and further experiments on ad keyword generation showed that our model has potential commercial value.
In the future, we plan to extend the contrastive module to more general settings, such as natural language understanding and representation learning.

\bibliographystyle{acl_natbib}
\bibliography{custom}

\end{document}